# Case-Factor Diagrams for Structured Probabilistic Modeling

**David McAllester**
TTI at Chicago
mcallester@tti-c.org

**Michael Collins**
CSAIL
Massachusetts Institute of Technology
mcollins@ai.mit.edu

**Fernando Pereira**
CIS
University of Pennsylvania
pereira@cis.upenn.edu

**Abstract**

We introduce a probabilistic formalism subsuming Markov random fields of bounded tree width and probabilistic context free grammars. Our models are based on a representation of Boolean formulas that we call case-factor diagrams (CFDs). CFDs are similar to binary decision diagrams (BDDs) but are concise for circuits of bounded tree width (unlike BDDs) and can concisely represent the set of parse trees over a given string under a given context free grammar (also unlike BDDs). A probabilistic model consists of a CFD defining a feasible set of Boolean assignments and a weight (or cost) for each individual Boolean variable. We give an inside-outside algorithm for simultaneously computing the marginal of each Boolean variable, and a Viterbi algorithm for finding the minimum cost variable assignment. Both algorithms run in time proportional to the size of the CFD.

## 1 Introduction

In this paper, we investigate efficient representations for structured probabilistic models. Informally, a structured model defines a distribution on structured objects such as sequences, parse trees, or assignments of values to variables. The number of possible structured objects typically grows exponentially in a natural measure of problem size. For example, the number of possible parse trees grows exponentially in the length of the string being parsed. Structured statistical models include Markov random fields (MRFs), probabilistic context free grammars (PCFGs), hidden Markov models (HMMs), conditional random fields (CRFs) [12], dynamic Bayes nets [11], probabilistic Horn abduction [15], and probabilistic relational models (PRMs) [10].

For each of these model types one can define a corresponding structured classification problem. In HMMs, for example, the problem is to recover the hidden state sequence from the observable sequence. For PCFGs, the problem is to recover a parse tree from a given word string. In PRMs, the problem is to recover latent entity labels and relations for a given set of observed entities and relations. We follow an approach where the statistical model defines $P(y|x)$ and structured classification finds the most likely $y$ for a given $x$. (Other approaches are possible – for example, maximum margin classifiers are discussed below.)

The structured statistical models discussed above are intuitively similar. They all involve local probability tables or local cost functions. It is widely believed that many, if not all, of the above modeling formalisms can be viewed as special cases of MRFs (undirected graphical models). More specifically, in a structured classification problem one should be able to represent $P(y|x)$ as an MRF. By assuming $P(y|x)$ is modeled as an MRF one can prove theorems and design algorithms and software at an abstract level which simultaneously applies to all of the modeling formalisms discussed above.

Unfortunately, for some of the above models the representation of $P(y|x)$ as an MRF is problematic. The most problematic case is perhaps PCFGs. It is fairly easy to construct an MRF representing $P(y|x)$ where $y$ is a parse tree and $x$ is a word string. Unfortunately, standard MRF algorithms take exponential time when applied to the natural MRF representation. This is a somewhat surprising outcome, given that there are well-known inference algorithms for PCFGs which run in cubic time in the length of the word string $x$.

This paper presents a modeling formalism which handles both MRFs of bounded tree width and PCFGs.

First we define a *linear Boolean model* (LBM). An LBM consists of three parts: a set of boolean variables; a formula defining a set of possible assignments to these variables (a "feasible set"); and an assignment of a weight to each variable. The weight for a complete variable assignment is then the sum of weights for those variables in the assignment that are true. The weight associated with a truth assignment can be written as a linear function of the bits in the assignment — hence the term "linear". We show how to encode both standard MRFs and PCFGs, as LBMs.

The main problem we solve is how to encode compactly the set of possible assignments to the variables in an LBM in a single formalism handling both MRFs of bounded tree width and PCFGs. The *case-factor diagrams* (CFDs) we introduce for that purpose are similar to binary-decision diagrams (BDDs) [4]. CFDs differ from BDDs in two ways. First, CFDs are similar to *zero-suppressed* BDDs (ZBDDs) [14]. ZBDDs are designed for sparse truth assignments — truth assignments where most of the Boolean variables are false. Sparseness is important for representing PCFGs. In addition to being zero-surpressed, CFDs have "factor nodes" which allow a concise representation of problems that factor into independent subproblems. Factoring is important for representing MRFs of bounded tree width. We describe algorithms for CFDs that compute partition functions under Gibbs distributions for $P(y|x)$, that select the maximum likelihood (Viterbi) structure, and an inside-outside algorithm for computing the marginal distributions of all of the Boolean variables. These algorithms all run in time linear in the number of nodes in the CFD. We demonstrate that PCFG models can be encoded in a CFD which has $O(n^3)$ size where $n$ is the length of the input string. We also show that MRFs of bounded tree width can be represented by a CFD with a polynommial number of nodes.

There are various lines of related work. A variant of BDDs for circuits of bounded tree width was introduced by McMillan [13]. Although McMillan's formalism is more elaborate, it turns out that simply extending BDDs with "and" nodes suffices for representing MRFs of bounded width. But representing PCFGs seems to require a zero-suppressed formalism. CFDs are related to the recursive conditioning algorithm introduced by Darwiche [6, 1]. The nodes of a CFD correspond to the "subproblems" that arise in recursive conditioning. Recursive conditioning cases on the value of a variable, factors the remaining problem into independent subproblems, and then solves the subproblems recursively. CFDs provide a data structure that explicitly represents this case-factor structure. The CFD allows different algorithms to exploit that same structure. CFDs and recursive conditioning can both exploit context-sensitive independence (CSI) [3]. CSI is particularly important for PCFGs where the tree width of the natural MRF representation is large and tractability is due to CSI. Darwiche [7] provides a representation of case-factor structure based on arithmetic expressions, but it is not clear how to represent the feasible set of parse trees in that form.. Dechter [8] presents a representation of case-factor structure based on and/or graphs, and also discusses representations in terms of BDDs, but she does not consider the question of sparse assignments which is critical for PCFGs. However, it is possible to represent the distribution of parse trees with an appropriate and/or graph. In doing this one must use the convention that the feasible set is in one-to-one correspondence with the "solution trees" of the graph. CFDs explicitly represent the feasible set as a set of truth assignments. In the case of PCFGs, the truth assignment representation of the feasible set is natural because each Boolean variable can be given a natural meaning as a statement about the parse tree represented by the assignment.

Developing a common language for structured modeling has potential applications to maximum-margin structured classification [16, 5, 2]. A maximum margin model is trained using an objective function stated in terms of margins rather than in terms of $P(y|x)$. However, the model parameters can still be viewed as defining an log-linear or maxent probabilistic representation. CFDs provide a formalism for structured modeling that allows these algorithms and others to be formulated at a level of generality that covers both MRFs of bounded tree width and weighted grammar formalisms like PCFGs.

## 2 Linear Boolean Models

We first fix some notation and terminology. Given a set of *variables* $V$ and *domains* dom($x$) for each $x \in V$, an *assignment* $\rho$ maps $x \in V$ to $\rho(x) \in$ dom($x$); a *partial* assignment $\sigma$ maps a subset of the variables dom($\sigma$) $\subseteq V$ to appropriate values. The number of variables given values by (partial) assignment $\rho$ is $|\rho| = |\text{dom}(\rho)|$. We write $\rho' \sqsubseteq \rho$ if dom($\rho'$) $\subseteq$ dom($\rho$) and $\rho'(x) = \rho(x) \; \forall x \in$ dom($\rho'$). If $\rho$ is a (possibly partial) assignment on $V$ and $V' \subseteq V$, $\rho|_{V'}$ is the unique assignment such that $\rho|_{V'} \sqsubseteq \rho$ and dom($\rho|_{V'}$) = dom($\rho$) $\cap V'$. If all the variables are *Boolean*, that is dom($x$) = $\mathbb{B}$ = $\{0,1\} \; \forall x \in V$, the assignment is a *truth* assignment. If $\rho$ is a (possibly partial) assignment, $x \in V$ a variable, and $v \in$ dom($x$), $\rho[x := v]$ is the assignment identical to $\rho$ except that $\rho[x := v](x) = v$. If $F$ is a set of assignments, $F[x := v] = \{\rho[x := v] : \rho \in F\}$. If $\rho$ and $\sigma$ are truth assignments, $\rho \vee \sigma$ is the assignment such that $(\rho \vee \sigma)(x) = 1$ if and only if $\rho(x) = 1$ or $\sigma(x) = 1$. If $F_1$ and $F_2$ are sets of truth assignments, $F_1 \vee F_2 = \{\rho \vee \sigma : \rho \in F_1 \text{ and } \sigma \in F_2\}$. The *support* of a truth assignment is the set of variables set to 1 by the assignment.

We can now describe a general class of structured probabilistic models with Boolean variables. A *linear Boolean model* (LBM) is a triple $\langle V, F, \Psi \rangle$ where $V$ is a set of Boolean variables, $F$ is a set of feasible configurations, each of which is a truth assignment to $V$, and $\Psi$ is an *energy function* $\Psi : V \to \mathbb{R}$. We extend $\Psi$ to configurations $\rho \in F$ with the following "linear" definition:

$$\Psi(\rho) = \sum_{z \in V} \Psi(z)\rho(z) \quad (1)$$

If we view $\Psi$ as a vector in $R^{|V|}$ and $\rho$ as a vector in $\mathbb{B}^{|V|}$ then $\Psi(\rho)$ is simply the inner product of $\Psi$ and $\rho$. A LBM $M$ defines a probability distribution $P(\cdot \mid F, \Psi)$ on feasible configurations $\rho \in F$ as follows.

$$P(\rho|F, \Psi) = \frac{1}{Z(F, \Psi)} e^{-\Psi(\rho)} \quad (2)$$

$$Z(F, \Psi) = \sum_{\rho \in F} e^{-\Psi(\rho)} \quad (3)$$

Given equation (2) we have that an LBM is really just a log-linear or maxent model [9] on a set $F$ under the restrictions that all features are Boolean and that each element of $F$ is uniquely determined by its feature values. A critical issue is how to represent the feasible set $F$. Before discussing the representation of $F$, however, we give two examples of representing structured models with LBMs.

## 3 Markov Random Fields

A Markov random field (MRF) consists of variables and energy terms on configurations of those variables. More precisely, we assume a finite set of variables $y_1$, ... $y_\ell$ with associated domains $\mathcal{Y}_1$, ..., $\mathcal{Y}_\ell$. We take the domains $\mathcal{Y}_j$ to be finite sets with $|\mathcal{Y}_i| \geq 2$. We define a *configuration* to be an assignment $\rho$ of values to the variables. An MRF is a set of such variables plus a set of energy terms $\Psi_1$, ..., $\Psi_N$ each of which maps a configuration to a real number. Any such set of energy terms defines a hypergraph on the variables. More specifically, we say that $\Psi_k$ depends on variable $y_j$ if there exists configurations $\rho$ and $\rho'$ which agree on all variables except $y_j$ and such that $\Psi_k(\rho) \neq \Psi_k(\rho')$. Let $V_k$ denote the set of variables on which $\Psi_k$ depends. The sets $V_k$ define a hypergraph on the variables. If $|V_k| = 2$ for all $k$ then these sets define a graph.

An MRF $M$ defines a probability distribution over configurations $P(\rho|M)$ by the following equations:

$$P(\rho|M) = \frac{1}{Z(M)} e^{-\Psi(\rho)}$$
$$Z(M) = \sum_\rho e^{-\Psi(\rho)}$$
$$\Psi(\rho) = \sum_k \Psi_k(\rho)$$

To represent an MRF as a LBM we must represent a configuration of $M$ as a truth assignment on Boolean variables and represent the energy terms by an energy function on Boolean variables. Given an MRF $M$ we construct Boolean variables of the form "$y_i = v$" with $y_i$ a variable of $M$ and with $v \in \mathcal{Y}_i$. For each energy term $\Psi_k$ with $V_k = \{y_1, \ldots, y_m\}$ and each tuple of values $v_1, \ldots, v_k$ with $v_i \in \mathcal{Y}_i$ we also introduce the Boolean variable "$k, y_1 = v_1 \wedge \cdots \wedge y_m = v_m$". Of course not all truth assignments to these Boolean variables correspond to configurations of the random field $M$. In order for a Boolean assignment to be feasible we must have that for each $y$ exactly one of "$y = v_1$", ..., "$y = v_n$" is true and furthermore "$k, y_1 = v_1 \wedge \cdots \wedge y_m = v_m$" is true if and only if each of "$y_1 = v_1$", ..., "$y_m = v_m$" is true. Section 5 discusses a method for representing this feasible set of truth assignments. Finally we define the variable energy function as follows.

$$\Psi("y = v") = 0$$
$$\Psi("k, y_1 = v_1 \wedge \cdots \wedge y_m = v_m") = \Psi_k(v_1, \ldots, v_m)$$

## 4 Parse Distributions as LBMs

A CFG in Chomsky normal form is a set of productions of the following form where $X$, $Y$ and $Z$ are nonterminal symbols and $a$ is a terminal symbol.

$$X \to YZ$$
$$X \to a$$

A parse tree is a tree each node of which is labeled by a production of the grammar in the standard way. In a weighted CFG each production $X \to \gamma$ is assigned an energy (weight) $\Psi(X \to \gamma)$. For any parse tree $y$ we write yield($y$) for the yield of $y$, i.e., the sequence of terminal symbols at the leaves of the parse tree. We write $\Psi(y)$ for the total energy of the parse tree $y$ — $\Psi(y)$ is the sum over all nodes of $y$ of the energy of the production used at that node. For a given string $x$ of terminal symbols we have a probability distribution on parse trees $y$ with $\text{yield}(y) = x$ defined as follows.

$$P(y|x) = \frac{1}{Z(x)} e^{-\Psi(y)} \quad (4)$$

$$Z(x) = \sum_{y:\ \text{yield}(y)=x} e^{-\Psi(y)} \quad (5)$$

To construct an LBM representation of $P(y|x)$ we first define a set of Boolean variables. Let $n$ be the length of $x$. First we have a phrase variable "$X_{i,j}$" for each nonterminal $X$ in the grammar and $1 \leq i < j \leq n+1$. This phrase variable represents the statement that the parse contains a phrase with nonterminal $X$ spanning the string from $i$ to $j-1$ inclusive. Second we have a branch variable "$X_{i,k} \to Y_{i,j} Z_{j,k}$" for each production $X \to YZ$ in the grammar and $1 \leq i < j < k \leq n+1$. A branch variable represents the statement that the parse contains a node labeled with the given production where the left child of the node spans the string from $i$ to $j-1$ and the right child spans $j$ to $k-1$. Finally, we have a terminal variable "$X_{i,i+1} \to a$" for each terminal production $X \to a$ and position $i$ in the input string. A terminal variable represents the statement that the parse tree produces terminal symbol $a$ from nonterminal $X$ at position $i$. We take $V$ to be the set of all such phrase, branch, and terminal variables. Each parse tree determines a truth assignment to the variables in $V$ and we take $F$ to be the set of assignments corresponding to parse trees. Finally, we must define the energy of each Boolean variable. The variable energy function $\Psi$ is given by the following equations.

$$\Psi(\text{"}X_{i,j}\text{"}) = 0$$
$$\Psi(\text{"}X_{i,k} \to Y_{i,j} Z_{j,k}\text{"}) = \Psi(X \to YZ)$$
$$\Psi(\text{"}X_{i,i+1} \to a\text{"}) = \Psi(X \to a)$$

## 5 Case Factor Diagrams (CFDs)

A *case-factor diagram* represents the feasible set by a search tree over the set of possible truth assignments. The search tree cases on the value of individual variables and factors the feasible set into a product of independent feasible sets when possible. We represent this case-factor search tree by an expression.

**Definition 1** *A case-factor diagram (CFD) $D$ is an expression generated by the following grammar where $x$ is a Boolean variable; a case expression $\text{case}(x, D_1, D_2)$ must satisfy the constraint that $x$ does not appear in $D_1$ or $D_2$; and a factor expression $\text{factor}(D_1, D_2)$ must satisfy the constraint that no variable occurs in both $D_1$ and $D_2$.*

$$D ::= \text{case}(x, D_1, D_2) \mid \text{factor}(D_1, D_2) \mid \text{unit} \mid \text{empty}$$

We denote by $V(D)$ the set of variables occurring in $D$.

To define the meaning of CFDs, it is convenient to see all CFD variables as members of a common countably infinite set of variables $V$. The interpretation $F(D)$ of a CFD $D$ is then a finite set of finite support assignments to $V$. We use $\overline{0}$ for the totally false assignment (the zero vector). $F(D)$ is defined as follows.

$$F(\text{unit}) = \{\overline{0}\}$$
$$F(\text{empty}) = \emptyset$$
$$F(\text{case}(x, D_1, D_2)) = F(D_1)[x := 1] \cup F(D_2)$$
$$F(\text{factor}(D_1, D_2)) = F(D_1) \vee F(D_2)$$

Therefore, like in ZBDDs, variables that are false in all assignments in $F(D)$ are not mentioned in $D$. In contrast, a BDD must test all variables in its domain, precluding the compact representation of sparse assignments.

An an example consider variables $x_1, x_2, \ldots$ and consider the CFD $A_i$ defined as follows.

$$A_0 = \text{unit}$$
$$A_{i+1} = \text{case}(x_{i+1}, A_i, A_i)$$

Under the semantics stated above we have that $F(A_i)$ is the set of all the $2^i$ truth assignments $\rho$ satisfying the constraint that $\rho(x_j) = 0$ for all $j > i$. As another example, consider $B_i$ defined as follows.

$$B_0 = \text{unit}$$
$$B_{i+1} = \text{factor}(\text{case}(x_{i+1}, \text{unit}, \text{unit}), B_i)$$

We leave it to the reader to verify that $F(B_i) = F(A_i)$. As a third example consider $C_i$ defined as follows.

$$C_0 = \text{unit}$$
$$C_{i+1} = \text{case}(x_{i+1}, C_i, \text{empty})$$

We have that $F(C_i)$ contains only the single truth assignment $\rho$ such that $\rho(x_j) = 1$ for $j \leq i$ and $\rho(x_j) = 0$ for $j > i$. In general this semantics has the property that if $x$ does not occur in $D$ then $\rho(x) = 0$ for any assignment $\rho \in F(D)$. Because the two arguments of a factor expression cannot share variables, we have that the number of assignments in $F(\text{factor}(D_1, D_2))$ equals the number of assignments in $F(D_1)$ times the number of assignments in $F(D_2)$. We leave it to the reader to verify that any feasible set on any finite set of variables can be represented by a CFD.

The meaning of CFD expressions is independent of their representation as data structures. However, the running time of algorithms depends crucially on that representation. For all the algorithms we discuss, we assume that CFD expressions are represented as *diagrams*, which are DAGs with one node for each distinct subexpression, and edges from the node for an expression to the nodes for its immediate subexpressions. That is, common subexpressions are represented uniquely. For example, the CFD $A_i$ defined above viewed as a tree has $2^i$ leaves. Viewed as a diagram, however, $A_i$ has only $i+1$ nodes but $2^i$ different paths from the root node to the leaf node. The size of a CFD $D$, denoted $|D|$, is defined to be the number of distinct subexpressions of $D$ (including $D$ itself). In other words, $|D|$ is the number of nodes in the diagram view of $D$. We will often use the word "node" as a synonym for "expression". We will also use the standard DAG notions of parent, child, and (directed) path for CFDs. We write $D' \preceq D$ to state that node $D'$ is a (possibly improper) descendant of node $D$. If $D' \preceq D$, the *depth* of $D'$ (in $D$) is the length of the longest path from $D$ to $D'$.

## 6 CFDs for MRFs

Here we define a CFD representation of the feasible set for the LBM constructed in section 3. Consider the problem of computing $Z(M)$ for an MRF $M$. We assume that the variables of $M$ have been given in a fixed order $y_1, y_2, \ldots, y_n$. The assignments to these variables form a tree whose root has a branch for each value of $y_1$, the next level branches for each value of $y_2$ and so on. As variables are assigned, however, the residual hypergraph defined by the energy terms often factors into disjoint sets of terms on disjoint sets of variables. So one can compute $Z(M)$ by factoring the residual problem when possible and, if no factoring is possible, casing out on the value of the next variable (after which more factoring may be possible). This "case-factor process" determines a set of subproblems. The nodes (subexpressions) in the CFD representation of the MRF correspond to the subproblems that arise in this way. Each such subproblem is defined by a subset $\Sigma$ of the energy terms and a partial assignment $\rho$ to (some of) the variables occurring in $\Sigma$.

More formally, consider a subset $\Sigma$ of the energy terms of $M$. Let $V(\Sigma)$ be the set of variables on which some energy term in $\Sigma$ depends, i.e., $V(\Sigma) = \cup_{k \in \Sigma} V_k$. Let $\rho$ be a partial assignment of values to (some of) the variables in $V(\Sigma)$. Note that $\rho$ is defined on the general variables of $M$ rather than the Boolean variables of $M'$. For each pair of such a subset $\Sigma$ and partial assignment $\rho$ we now define a CFD $D(\Sigma, \rho)$. The CFD for the full feasible constraint is $D(\Sigma(M), \emptyset)$ where $\Sigma(M)$ is the set of all energy terms in $M$ and $\emptyset$ is the empty partial assignment. For a given partial assignment $\rho$ we define a graph structure on the energy terms in $\Sigma$ by saying that there is an edge between two energy terms if there is a variable not assigned a value by $\rho$ on which both terms depend. The key to concise representation is to factor the problem when $\Sigma$ becomes disconnected. We use the notation $\mathsf{case}(\langle z_1, D_1 \rangle, \langle z_2, D_2 \rangle, \ldots, \langle z_m, D_m \rangle)$ as an abbreviation for $\mathsf{case}(z_1, D_1, \mathsf{case}(\langle z_2, D_2 \rangle, \ldots, \langle z_n, D_n \rangle))$ where $\mathsf{case}(\langle z, D \rangle)$ is $\mathsf{case}(z, D, \mathsf{empty})$. The CFD $D(\Sigma, \rho)$ is defined as follows.

1. If $\Sigma$ is disconnected under partial assignment $\rho$, let $\Sigma = \Sigma_1 \cup \Sigma_2$ where $\Sigma_1$ and $\Sigma_2$ are disjoint and not connected to each other. Then:

    $D(\Sigma, \rho) = \mathsf{factor}(D(\Sigma_1, \rho|_{V(\Sigma_1)}), D(\Sigma_2, \rho|_{V(\Sigma_2)}))$

2. Otherwise, if $\Sigma$ consists of a single constraint $\Psi_k$ and $\rho$ assigns values to all of $V(\Sigma)$, we have the following where $V_k = \{y_1, \ldots, y_m\}$ and $v_i = \rho(y_i)$.

    $D(\Sigma, \rho) =$
    $\mathsf{case}(\texttt{"}k, y_1 = v_1, \ldots, y_m = v_m\texttt{"}, \mathsf{unit}, \mathsf{empty})$

3. Otherwise, let $y$ be the earliest variable (under the given variable order) in $V(\Sigma)$ that is not in $\mathrm{dom}(\rho)$. In this case we have the following where $\mathrm{dom}(y) = \{v_1, \ldots, v_n\}$.

    $D(\Sigma, \rho) =$
    $\mathsf{case} \begin{pmatrix} \langle \texttt{"}y = v_1\texttt{"}, D(\Sigma, \rho[y := v_1]) \rangle, \\ \vdots \\ \langle \texttt{"}y = v_n\texttt{"}, D(\Sigma, \rho[y := v_n]) \rangle \end{pmatrix}$

We now show that MRFs of small tree width have concise CFD representations. First we define the notion of tree width.

**Definition 2** *We consider a fixed variable order $y_1 \ldots y_n$. For $i$ with $1 \leq i \leq n$ we define the present variable to be $y_i$, the past variables to be all variables $y_j$ with $j \leq i$, and the future variables to be all variables $y_j$ with $j \geq i$. Note that the present variable is both past and future. We define $G_i$ to be the graph whose nodes are the energy terms of $M$ and where two energy terms are connected by an edge if they both depend on the same future variable. The connected components of $G_i$ give independent subproblems on the future variables. If $\Sigma$ is a connected component of $G_i$ then we define the width of $\Sigma$ (at time $i$) to be the number of past variables occurring in $\Sigma$. The tree width of $M$ under the given variable ordering is the maximum over all $i$ of the maximum width (at time $i$) of a connected component of $G_i$.*

We now have the following theorem.

**Theorem 1** *Let $w$ be the tree width of $M$ under the given variable ordering. Then $|D(M)|$ is $O(Nd^w)$ where $N$ is the number of energy terms in $M$ and $d = \max_i |\mathcal{Y}_i|$.*

**Proof:** We first show that the total number of nodes of the CFD can be no more than twice the number of pairs $\langle \Sigma, \rho \rangle$ where $\Sigma$ is a connected component of $G_i$ for some $i$ and $\rho$ is a partial assignment to past variables of $\Sigma$. All nodes in the CFD are either of this form, are on of the constants unit, or empty, or are factor nodes generated by step 1 of the procedure. Suppose the top level problem can be factored into some number of independent subproblems. The factoring is represented by a binary tree whose leaves are the final factors, so the number of nodes in the tree is proportional to the number of factors. A similar observation applies to any factoring that occurs following a case analysis introduced by step 3. So, without loss of generality, we need only consider pairs $\langle \Sigma, \rho \rangle$ where $\Sigma$ is a connected component of $G_i$. The set of all such subsets forms a tree whose leaves consist of single energy terms. Hence there are at most $2N$ such subsets. For a fixed subset $\Sigma$, the set of possible assignments to past variables form a tree with at most $d^w$ leaves. So there are at most $2d^w$ values of $\rho$ for a given value of $\Sigma$. ∎

## 7 CFDs for Parsing

Here we construct a CFD for the feasible set of the LBM defined in Section 4 for a grammar $G$. We define the CFD $D(\texttt{"}X_{i,k}\texttt{"})$ such that the assignments in $F(D(\texttt{"}X_{i,k}\texttt{"}))$ are in one-to-one correspondence with the parse trees of the span from $i$ to $k-1$ with root nonterminal $X$. The CFD representing the full feasible set of parses is $D(\texttt{"}S_{1,n+1}\texttt{"})$. First we define $D(\texttt{"}X_{i,k}\texttt{"})$ as follows where $B(\texttt{"}X_{i,k}\texttt{"})$ represents the consequences of making $\texttt{"}X_{i,k}\texttt{"}$ true.

$$D(\texttt{"}X_{i,k}\texttt{"}) = \mathsf{case}(\texttt{"}X_{i,k}\texttt{"}, B(\texttt{"}X_{i,k}\texttt{"}), \mathsf{empty})$$

For $k > i+1$ we define the consequences $B(\texttt{"}X_{i,k}\texttt{"})$ as follows using the multi-branch case notation defined in section 6.

$$B(\texttt{"}X_{i,k}\texttt{"}) = \mathsf{case}(\langle b_1, \ B(b_1)\rangle, \ldots, \langle b_n, \ B(b_n)\rangle)$$

where the variables $b_p$ are all possible branch variables of the form $\texttt{"}X_{i,k} \to Y_{i,j} Z_{j,k}\texttt{"}$, and $B(\texttt{"}X_{i,k} \to Y_{i,j}Z_{j,k}\texttt{"}) = \mathsf{factor}(D(\texttt{"}Y_{i,j}\texttt{"}), D(\texttt{"}Z_{j,k}\texttt{"}))$.

Finally, if $a_i$ is the $i$th input symbol, we have

$$B(\texttt{"}X_{i,i+1}\texttt{"}) = \begin{cases} \mathsf{case}(\texttt{"}X_{i,i+1} \to a_i\texttt{"}, \mathsf{unit}, \mathsf{empty}) & \text{if } X \to a_i \in G \\ \mathsf{empty} & \text{otherwise} \end{cases}$$

This construction has the property that $|D(\texttt{"}S_{1,n+1}\texttt{"})|$ is $O(|G|n^3)$ where $|G|$ is the number of productions in the grammar.

## 8 Inference on CFD Models

A *CFD model* $\langle D, \Psi \rangle$ is an LBM whose feasible set is defined by a CFD $D$ and whose energy function $\Psi$ assigns costs to the variables of $D$. We will now present the main inference algorithms on CFDs.

**The Inside Algorithm.** We first consider the problem of computing $Z(F(D), \Psi)$ as defined by equation (3). Here we write $Z(D, \Psi)$ as an abbreviated form of $Z(F(D), \Psi)$. It turns out that $Z(D, \Psi)$ can be computed by recursive descent on subexpressions of $D$ using the following equations.

$$Z(\mathsf{case}(x, D_1, D_2), \Psi) = e^{-\Psi(x)} Z(D_1, \Psi) + Z(D_2, \Psi)$$
$$Z(\mathsf{factor}(D_1, D_2), \Psi) = Z(D_1, \Psi) Z(D_2, \Psi)$$
$$Z(\mathsf{unit}, \ \Psi) = 1$$
$$Z(\mathsf{empty}, \ \Psi) = 0$$

The correctness of these equations can be proved by induction on the size of $D$. By caching these computations for each subexpression of $D$, these equations give a way of computing $Z(D, \Psi)$ in time proportional to $|D|$. These equations are analogous to the inside algorithm used in statistical parsing.

**The Viterbi Algorithm.** Next we consider the problem of computing minimum energy over the elements of $F(D)$. In particular we define $\Psi^*(D, \Psi)$ as follows.

$$\Psi^*(D, \Psi) = \min_{\rho \in F(D)} \Psi(\rho)$$

We can compute $\Psi^*(D, \Psi)$ using the following equations.

$$\Psi^*(\mathsf{case}(z, D_1, D_2), \Psi) = \min \begin{pmatrix} \Psi(z) + \Psi^*(D_1, \Psi), \\ \Psi^*(D_2, \Psi) \end{pmatrix}$$
$$\Psi^*(\mathsf{factor}(D_1, D_2), \Psi) = \Psi^*(D_1, \Psi) + \Psi^*(D_2, \Psi)$$
$$\Psi^*(\mathsf{unit}, \Psi) = 0$$
$$\Psi^*(\mathsf{empty}, \Psi) = +\infty$$

Again the correctness of these equations can be proved by a direct induction on the size of $D$. These equations can easily be modified to also compute a truth assignment that achieves the minimum energy. This is a truth assignment of highest probability.

**Marginals.** Next we consider the problem of computing marginal probabilities of the form $P(z = 1 \mid D, \Psi, \sigma)$ where $\sigma$ is a partial truth assignment that fixes the values of some of the CFD model variables. We will show that these marginals can be computed in time proportional to $|D||\sigma|$.

The marginal $P(z = 1 \mid D, \Psi, \sigma)$ can be written as follows:

$$P(z \mid D, \Psi, \sigma) = \frac{Z(D, \Psi, \sigma[z := 1])}{Z(D, \Psi, \sigma)}$$

$$Z(D, \Psi, \sigma) = \sum_{\rho \in F(D): \sigma \sqsubseteq \rho} e^{-\Psi(\rho)}$$

So it suffices to be able to compute $Z(D, \Psi, \sigma)$. We now define the auxiliary quantity $Z'(D, \Psi, \sigma) = Z(D, \Psi, \sigma|_{V(D)})$. Our procedure computes $Z(D, \Psi, \sigma)$ by computing $Z'(D', \Psi, \sigma)$ for all subnodes $D'$ of $D$. Note that the number of such values is $|D|$. The $Z'$ values satisfy the following equations for factor, unit and empty expressions.

$$Z'(\mathsf{factor}(D_1, D_2), \Psi, \sigma) = Z'(D_1, \Psi, \sigma) Z'(D_2, \Psi, \sigma)$$
$$Z'(\mathsf{unit}, \Psi, \sigma) = 1$$
$$Z'(\mathsf{empty}, \Psi, \sigma) = 0$$

Computing $Z'$ on case expressions is more subtle. Let $D$ be the expression $\mathsf{case}(z, D_1, D_2)$. We now have the following equation where $Z(v, D, D', \Psi, \sigma)$ is defined below.

$$Z'(D, \Psi, \sigma) = \begin{cases} e^{-\Psi(z)} Z(z, D, D_1, \Psi, \sigma) & \text{if } \sigma(z) = 1 \\ Z(z, D, D_2, \Psi, \sigma) & \text{if } \sigma(z) = 0 \\ e^{-\Psi(z)} Z(z, D, D_1, \Psi, \sigma) \\ + Z(z, D, D_2, \Psi, \sigma) & \text{otherwise} \end{cases}$$

$Z(z, D, D', \Psi, \sigma)$ expresses the constraint that omitted variables default to 0 in CFDs. If $\sigma(z') = 1$ where $z'$ occurs in $D$ but not in $D'$, $Z(z, D, D', \Psi, \sigma) = 0$, otherwise $Z(z, D, D', \Psi, \sigma) = Z'(D', \Psi, \sigma)$.

To analyze the running time of computing $Z(D, \Psi, \sigma)$ we first note that there are a linear number of values needed of the form $Z(z, D', D'', \Psi, \sigma)$. Assuming unit time hash table operations, it is possible to cache the answer to all queries of the form $z \in D'$, for $z' \in \mathrm{dom}(\Sigma)$ and $D'$ a node in $D$, in $O(|D||\sigma|)$ time. Given this cache, each call to $Z(z, D, D', \Psi, \sigma)$ can be computed in time proportional to $|\sigma|$. So the overall computation takes time proportional to $|D||\sigma|$.

**The Inside-Outside Algorithm.** Using the above conditional probability algorithm to compute $P(z = 1 \mid D, \Psi)$ for all variables $z$ can take $\Omega(|D|^2)$ time. However, a generalization of the inside-outside algorithm can be used to simultaneously compute $P(z = 1 \mid D, \Psi)$ for all variables $z$ in $D$ in $O(|D|)$ time. The value $Z(D, \Psi)$ is the "inside" value associated with $D$. Intuitively, the outside value of a node in a CFD is the total weight of the "contexts" in which that node appears. Formalizing the appropriate notion of context for general CFDs is somewhat subtle and is done in the appendix. Although the definition of context is subtle, the equations for computing outside values are rather natural.

A node is *open* if it does contain variables and closed otherwise. Outside values are only defined for open nodes. For $D' \preceq D$, and $D'$ open, we now define the *outside value* $O(D', D, \Psi)$ of $D'$ (in $D$). For $D' = D$, $O(D, D, \Psi) = 1$. For $D' \neq D$ and $D'$ open, we have:

$$\begin{aligned} O(D', D, \Psi) &= \sum_{\mathsf{case}(z, D', D'') \preceq D} O(\mathsf{case}(z, D'D''), D, \Psi) e^{-\Psi(z)} \\ &+ \sum_{\mathsf{case}(z, D'', D') \preceq D} O(\mathsf{case}(z, D'', D'), D, \Psi) \\ &+ \sum_{\mathsf{factor}(D', D'') \preceq D} O(\mathsf{factor}(D', D''), D, \Psi) Z(D'', \Psi) \\ &+ \sum_{\mathsf{factor}(D'', D') \preceq D} O(\mathsf{factor}(D'', D'), D, \Psi) Z(D'', \Psi) \end{aligned} \quad (6)$$

Once the inside value of every node has been computed, these equations allows the outside values of open nodes to be computed from the top down. This top-down calculation can be done in time proportional to the number of nodes. Finally we can compute $P(z = 1|D, \Psi)$ as follows.

**Theorem 2**

$$P(z = 1|D, \Psi) = \frac{Z(D, \Psi, \emptyset[z := 1])}{Z(D, \Psi)}$$

$$Z(D, \Psi, \emptyset[z := 1]) = \sum_{\mathsf{case}(z, D', D'') \preceq D} \begin{pmatrix} O(\mathsf{case}(z, D'D''), D, \Psi) \\ e^{-\Psi(z)} \\ Z(D', \Psi) \end{pmatrix}$$

The proof, which requires a careful definition of the meaning of the outside numbers, is given in the appendix.

## 9 Conclusions

We have described a class of structured probabilistic models based on case-factor diagrams. We have also shown that for a given a weighted context free grammar $G$ and input string $x$ the conditional probability $P(y|x)$ can be represented by a CFD model with $O(|G|n^3)$ nodes. We have also shown that any MRF with tree width $w$ in which variables have $V$ possible values and with $N$ energy terms can be represented by a CFD model with $O(NV^w)$ nodes. We have shown that for an arbitrary CFD model, computing the partition function, most likely variable assignment, and the probability of each Boolean variable, can all be done in time linear in the number of nodes. We believe that CFD models will provide a common language for specifying algorithms and stating theorems that can play for structured probabilistic models a similar role to that of BDDs in Boolean inference problems.

**Acknowledgments** Michael Collins and Fernando Pereira were supported in this work by the National Science Foundation under grants 0347631 and EIA-0205456, respectively.

## Appendix: Proof of Theorem 2

The proof uses a series of lemmas on the meaning of outside values. For a PCFG, the outside value of a phrase is the total weight of all extensions of the phrase to a full parse tree. We will see that the corresponding notion for a CFD model $\langle D, \Psi \rangle$ is the outside value $O(D', D, \Psi)$ of a node $D' \preceq D$, with the contexts being certain assignments that "lead to" $D'$. More precisely, the set $\gamma(D, \rho)$ of nodes that an assignment $\rho$ to the variables of $D$ leads to is given by:

$$\gamma(D, \rho) = \{D\} \cup \gamma'(D, \rho)$$

$$\gamma'(\mathsf{case}(z, D_1, D_2), \rho) = \begin{cases} \gamma(D_1, \rho) & \text{if } \rho(z) = 1 \\ \gamma(D_2, \rho) & \text{if } \rho(z) = 0 \end{cases}$$

$$\gamma'(\mathsf{factor}(D_1, D_2), \rho) = \gamma(D_1, \rho) \cup \gamma(D_2, \rho)$$

$$\gamma'(\mathsf{unit}, \rho) = \{\mathsf{unit}\}$$

$$\gamma'(\mathsf{empty}, \rho) = \{\mathsf{empty}\}$$

For $D' \preceq D$ we define the set of assignments that lead to $D'$ as follows:

$$F(D', D) = \{\rho \in F(D) : \ D' \in \gamma(D, \rho)\}$$

Each assignment $\rho \in F(D', D)$ can be split into an *outside part* and an *inside part* $\rho = \rho{\uparrow}D' \vee \rho{\downarrow}D'$, where $\rho{\uparrow}D' = \rho|_{\mathrm{dom}(\rho) - V(D')}$ and $\rho{\downarrow}D' = \rho|_{V(D')}$. The set of contexts for $D'$ is then

$$O(D', D) = \{\rho{\uparrow}D' : \ \rho \in F(D', D)\}$$

We now prove the expected relationship between outside values and contexts:

**Lemma 3** *For $D' \preceq D$ with $D'$ open*

$$O(D', D, \Psi) = \sum_{\sigma \in O(D', D)} e^{-\Psi(\sigma)} \qquad (7)$$

The proof requires several auxiliary lemmas giving properties of $\gamma(D, \rho)$ and $F(D', D)$.

**Lemma 4** *If $D'$ is open and $D' \in \gamma(D, \rho)$ then $\gamma(D, \rho)$ contains exactly one path from $D$ to $D'$.*

**Proof:** The proof is by induction on $|D|$. The lemma is vacuously true if $D$ is closed. Now suppose $D$ is of the form $\mathsf{case}(z, U, W)$ and assume the lemma for $U$ and $W$. If $\rho(z) = 1$ then $\gamma(D, \rho)$ is $\{D\} \cup \gamma(U, \rho)$ and the lemma follows from the induction hypothesis on paths into $U$. A similar observation holds for paths into $W$ when $\rho(z) = 0$. Finally, suppose $D$ is $\mathsf{factor}(U, W)$. Since $U$ and $W$ do not share variables, any open node in $U$ or $W$ must be in only one of these nodes. The lemma again follows from the induction hypothesis on paths into $U$ and paths into $W$. ∎

**Lemma 5** *If $\rho \in F(D)$ then $\mathsf{empty} \notin \gamma(D, \rho)$.*

**Proof:** The proof is by induction on $|D|$. If $D = \mathsf{empty}$ then $F(D) = \emptyset$ and the lemma is vacuously true. If $D = \mathsf{unit}$, $\gamma(D, \rho) = \{\mathsf{unit}\}$, satisfying the lemma. Now suppose $D = \mathsf{case}(z, D', D'')$ and assume the lemma for $D'$ and $D''$. If $\rho(z) = 1$ then $\gamma(D, \rho) = \{D\} \cup \gamma(D', \rho)$. But by construction $\gamma(D', \rho)$ depends only on the values of $\rho$ on $V(()D')$, so $\gamma(D', \rho) = \gamma(D', \rho \downarrow D')$. By definition of $F(D)$, $\rho \downarrow D' \in F(D')$. So by the induction hypothesis $\mathsf{empty} \notin \gamma(D', \rho \downarrow D')$ and hence $\mathsf{empty} \notin \gamma(D, \rho)$. Similar arguments prove the cases $\rho(z) = 0$ and $D = \mathsf{factor}(D', D'')$. ∎

**Lemma 6** *If $\rho \in F(D', D)$ then $\rho \downarrow D' \in F(D')$.*

**Proof:** If $\rho \in F(D', D)$ and $D'$ is closed, by definition of $F(D', D)$ and of $\gamma(D, \rho)$, all subnodes of $D'$ are in $\gamma(D, \rho)$. By Lemma 5, none of those subnodes can be $\mathsf{empty}$. Therefore, $D'$ involves only $\mathsf{factor}$ and $\mathsf{unit}$, so $F(D') = \{\overline{0}\}$ and $\rho \downarrow D' = \overline{0}$ so the lemma follows. For open nodes $D'$ the proof is by induction on the depth of $D'$ in $D$. If $D' = D$, $F(D', D) = F(D)$ and the result is immediate. Now assume the lemma for all open nodes of depth $k$ in $D$ and consider an open node $D'$ in $D$ of depth $k+1$. Consider $\rho \in F(D', D)$. The set $\gamma(D, \rho)$ includes $D'$ and, by Lemma 4, a unique path from $D$ to $D'$. Let $D''$ be the parent of $D'$ in this path. By the induction hypothesis, $\sigma = \rho \downarrow D'' \in F(D'')$. In addition, by construction $D' \in \gamma(D'', \sigma)$ and hence $\sigma \in F(D', D'')$. To complete the proof, we just need to show that $\sigma \downarrow D' \in F(D')$. First suppose the parent $D''$ is of the form $\mathsf{case}(z, D', V)$ where $\sigma(z) = 1$. Then the result follows from the definition of $F(\mathsf{case}(z, D', V))$. Similar arguments apply when $D''$ is of one of the forms $\mathsf{case}(z, V, D')$, $\mathsf{factor}(D', V)$ or $\mathsf{factor}(V, D')$. ∎

**Lemma 7** *If $\sigma$ and $\sigma'$ agree on all variables not occurring in $D'$ then $D' \in \gamma(D, \sigma)$ if and only if $D' \in \gamma(D, \sigma')$.*

**Proof:** If $D'$ is closed then $\sigma' = \sigma$ and the result is immediate. For open nodes the proof is by induction on the depth of $D'$. If $D' = D$ then the result is immediate. Now assume the result for all open nodes of depth $k$ and consider an open node $D'$ of depth $k+1$. Let $\sigma$ and $\sigma'$ be two assignments that agree on all variables not in $D'$. Suppose $D' \in \gamma(D, \sigma)$. Let $D''$ the parent of $D'$ in the unique path from $D$ to $D'$ contained in $\gamma(D, \sigma)$. By the induction hypothesis, $D''$ is also included in $\gamma(D, \sigma')$. Now suppose $D''$ is of the form $\mathsf{case}(z, D', V)$ and $\sigma(z) = 1$. Since $z$ does not occur in $D'$ we have $\sigma'(z) = 1$ and hence $D' \in \gamma(D, \sigma')$. A similar argument holds if $D'$ is of the form $\mathsf{case}(z, V, D')$, $\mathsf{factor}(D', V)$ or $\mathsf{factor}(V, D')$. The converse where we assume $D' \in \gamma(D, \sigma')$ is similar. ∎

**Lemma 8** *If $\sigma \in O(D', D)$ and $\rho \in F(D')$ then $\sigma \vee \rho \in F(D', D)$.*

**Proof:** If $D'$ is closed then $\rho = \overline{0}$ and $\sigma \in F(D)$ and the result is immediate. For open nodes the proof is by induction on the depth of $D'$. For $D' = D$, $O(D', D) = \{\overline{0}\}$ and $F(D', D) = F(D)$, so the result is immediate. Now assume the result for all open nodes of depth $k$ in $D$ and consider an open node $D'$ depth $k+1$. Consider $\sigma \in O(D', D)$ and $\rho \in F(D')$. By definition of $O(D', D)$, there is $\sigma' \in F(D', D)$ such that $\sigma' \uparrow D' = \sigma$. We must show that $\sigma' \uparrow D' \vee \rho \in F(D', D)$. Since $D' \in \gamma(D, \sigma')$, Lemma 7 implies $D' \in \gamma(D, \sigma' \uparrow D' \vee \rho)$. It remains only to show that $\sigma' \uparrow D' \vee \rho \in F(D)$. Let $D''$ be the unique parent of $D'$ in $\gamma(D, \sigma')$. Suppose $D''$ is of the form $\mathsf{case}(z, D', V)$ with $\rho(z) = 1$. We now apply the induction hypothesis to the pair of assignments $\sigma' \uparrow D''$ and $\rho[z := 1]$ to conclude that $\sigma' \uparrow D'' \vee \rho[z := 1] \in F(D)$. But $\sigma' \uparrow D'' \vee \rho[z := 1] = \sigma' \uparrow D' \vee \rho$, therefore and hence $\sigma' \uparrow D' \vee \rho \in F(D)$. A similar argument holds if $D'' = Case(z, V, D')$ with $\rho(z) = 0$. Now suppose $D'' = \mathsf{factor}(D', V)$. By Lemma 6, $\sigma' \downarrow V \in F(V)$. By the definition of $F(\mathsf{factor}(D', V))$, $\rho \vee \sigma' \downarrow V \in F(\mathsf{factor}(D', V))$. We now apply the induction hypothesis to $\sigma' \uparrow D'$ and $\rho \vee \sigma' \downarrow V$ to conclude $\sigma' \uparrow D' \vee \rho \vee \sigma' \downarrow V \in F(D)$. But $\sigma' \uparrow D' \vee \rho \vee \sigma' \downarrow V = \sigma' \uparrow D' \vee \rho$, thus $\sigma' \uparrow D' \vee \rho \in F(D)$. The case $D'' = \mathsf{factor}(V, D')$ is similar. ∎

Now for $\sigma \in O(D', D)$ we define $\gamma(D', D, \sigma)$ to be the unique path from $D$ to $D'$ contained in $\gamma(D, \sigma \vee \rho)$ for arbitrary $\rho \in F(D')$.

**Lemma 9** *If $\sigma \in O(D', D)$, $\gamma(D', D, \sigma)$ is well defined.*

**Proof:** By Lemma 8, $\sigma \vee \rho \in F(D', D)$ for any $\rho \in F(D')$, and by Lemma 4 the path $p$ from $D$ to $D'$ in $\gamma(D, \sigma \vee \rho)$ is unique. Now let $\rho' \in F(D')$, and let $D''$ be any node on the path from $D$ to $D'$ in $\gamma(D, \sigma \vee \rho)$. By Lemma 7, $D'' \in \gamma(D, \sigma \vee \rho')$. Hence $\gamma(D, \sigma \vee \rho')$ must contain $p$. ∎

**Lemma 10** *For $D' \preceq D$, $F(D', D) = \{\sigma \vee \rho : \sigma \in O(D', D) \text{ and } \rho \in F(D')\}$.*

**Proof:** For $\rho \in F(D', D)$, $\rho = \rho{\uparrow}D' \vee \rho{\downarrow}D'$. By the definition of $O(D', D)$, $\rho{\uparrow}D' \in O(D', D)$, and by Lemma 6, $\rho{\downarrow}D' \in F(D')$, so $\rho$ has the desired form. The converse follows from Lemma 8. ∎

**Proof of Lemma 3:** The proof is by induction on depth of $D'$ in $D$. If $D' = D$, by definition $O(D', D, \Psi) = 1$ and $O(D, D) = \{\overline{0}\}$, so the result follows. Now assume that the lemma holds for nodes of depth $k$ or less and let $D'$ be a node of depth $k + 1$. By the induction hypothesis, each occurrence of $O(\cdot, D, \Psi)$ in the right-hand side of (6) satisfies the lemma. By Lemma 9, $O(D', D)$ can be split into four disjoint subsets corresponding to the four terms on the right-hand side of (6). We must show that each of these terms computes an appropriate sum over an appropriate subset of $O(D', D)$. Consider the third term of the sum. This term corresponds to the set of assignments $\sigma = O(D', D)$ such that the parent of $D'$ in the path $\gamma(D', D, \sigma)$ is of the form $\mathsf{factor}(D', D'')$. Now consider a fixed parent $P$ of this form, and let $O = \{\sigma \in O(D', D) : P \in \gamma(D', D, \sigma)\}$. We must show that

$$\sum_{\sigma \in O} e^{-\Psi(\sigma)} = O(P, D, \Psi) Z(D'', \Psi) \qquad (8)$$

By the induction hypothesis, $O(P, D, \Psi) = \sum_{\sigma' \in O(P, D)} e^{-\Psi(\sigma')}$. By definition, $Z(D'', \Psi) = \sum_{\rho \in F(D'')} e^{-\Psi(\rho)}$. Therefore, the equality (8) holds if $O = \{\sigma' \vee \rho : \sigma' \in O(P, D) \text{ and } \rho \in F(D'')\}$. If $\sigma \in O$, it is easy to see that there are $\sigma' \in O(P, D)$ and $\rho \in F(D'', D)$ such that $\sigma = \sigma' \vee \rho$. By Lemma 6, $\rho \in F(D'')$. Conversely, consider $\sigma \in O(P, D)$ and $\rho \in F(D'')$. Consider any $\rho' \in F(D')$. The assignment $\rho \vee \rho'$ is in $F(P)$. By Lemma 8 we then have that $\sigma \vee \rho \vee \rho' \in F(D', D)$. But this implies that $\sigma \vee \rho \in O(D', D)$. This proves (8) and hence that the third term in the definition of $O(D', D, \Psi)$ has the appropriate semantics. The proof of the appropriate semantics for the other terms is similar. ∎

**Lemma 11** *If $\rho \in F(D)$ and $\rho(z) = 1$ then $\gamma(D, \rho)$ contains a node of the form $\mathsf{case}(z, D_1, D_2)$.*

**Proof:** The proof is by induction on $|D|$. If $D = \mathsf{empty}$, $F(D) = \emptyset$ and the lemma is vacuously true. If $D = \mathsf{unit}$, then $\rho = \overline{0}$ and there is no $z$ with $\rho(z) = 1$. If $D = \mathsf{factor}(D', D'')$ then either $(\rho{\downarrow}D')(z) = 1$ or $(\rho{\downarrow}D'')(z) = 1$ and the lemma follows form the induction hypothesis on $D'$ or $D''$. Finally, suppose $D = \mathsf{case}(w, D', D'')$. If $w = z$ then $D$ is of the desired form. Now assume $w \neq z$. If $\rho(w) = 1$ then we have $\rho[w := 0] \in F(D')$, $(\rho[w := 0])(z) = 1$, and $\gamma(D', \rho[w := 0]) \subseteq \gamma(D, \rho)$. In this case the lemma follows from the induction hypothesis. A similar argument holds for $\rho(w) = 0$. ∎

**Lemma 12** *For a given variable $z$, the set $\gamma(D, \rho)$ contains at most one node of the form $\mathsf{case}(z, D', D'')$.*

**Proof:** The proof is by induction on $|D|$. The lemma is immediate if $D$ is closed. Now suppose $D = \mathsf{case}(w, D', D'')$. If $w = z$ then $D = \mathsf{case}(z, D', D'')$ and there can be no other node of this form because $D'$ and $D''$ cannot contain $z$. Now suppose $w \neq z$. If $\rho(w) = 1$ then $\gamma(D, \rho)$ is $\{D\} \cup \gamma(D', \rho)$ and the lemma follows from the induction hypothesis on $D'$. A similar arguments holds if $\rho(w) = 0$. Finally, suppose $D$ is $\mathsf{factor}(D', D'')$. In this case the lemma follows from the induction hypothesis on $D'$ and $D''$ and the fact that $z$ can occur in at most one of $D'$ and $D''$. ∎

**Proof of theorem 2:** Let $F(z, D) = \{\rho \in F(D) : \rho(z) = 1\}$. For any $\rho \in F(z, D)$, by lemmas 11 and 12, there is a unique node $C(z, D, \rho) \in \gamma(D, \rho)$ of the form $\mathsf{case}(z, D_1, D_2)$. Let

$$O(z, D, D_1, D_2) = \\ \{\rho \in F(z, D) : C(z, D, \rho) = \mathsf{case}(z, D_1, D_2)\}$$

Then

$$Z(D, \Psi, \emptyset[z := 1])$$
$$= \sum_{\rho \in F(z, D)} e^{-\Psi(\rho)}$$
$$= \sum_{\mathsf{case}(z, D_1, D_2) \preceq D} \sum_{\rho \in O(z, D, D_1, D_2)} e^{-\Psi(\rho)}$$

The theorem thus follows if

$$\sum_{\rho \in O(z, D, D_1, D_2)} e^{-\Psi(\rho)} = \\ O(\mathsf{case}(z, D_1, D_2), \Psi) e^{-\Psi(z)} Z(D_1, \Psi)$$

This is true if $O(z, D, D_1, D_2)$ is the set of assignments of the form $\sigma \vee \emptyset[z := 1] \vee \rho$ with $\sigma \in O(\mathsf{case}(z, D_1, D_2), D)$ and $\rho$ in $F(D_1)$, which follows by an argument similar to that used for Lemma 3. ∎